\documentclass[10pt,twocolumn,letterpaper]{article}
\usepackage[pagenumbers]{cvpr}
\usepackage{graphicx}
\usepackage{amsmath}
\usepackage{amssymb}
\usepackage{booktabs}
\usepackage{physics}
\usepackage[norelsize, linesnumbered, ruled, lined, boxed, commentsnumbered]{algorithm2e}
\usepackage{placeins}
\usepackage[pagebackref,breaklinks,colorlinks]{hyperref}
\usepackage[capitalize]{cleveref}
\crefname{section}{Sec.}{Secs.}
\Crefname{section}{Section}{Sections}
\Crefname{table}{Table}{Tables}
\crefname{table}{Tab.}{Tabs.}

\newcommand{\name}{CLIP2StyleGAN\xspace}

\DeclareMathOperator*{\argmin}{arg\,min}

\def\cvprPaperID{***} 
\def\confName{CVPR}
\def\confYear{2022}

\begin{document}

\title{\name: Unsupervised Extraction of StyleGAN Edit Directions}
\author{Rameen Abdal\textsuperscript{1} \quad Peihao Zhu\textsuperscript{1} \quad John Femiani\textsuperscript{2} \quad Niloy J.  Mitra\textsuperscript{3}
\quad Peter Wonka\textsuperscript{1} \\
\\
\textsuperscript{1}KAUST \quad \textsuperscript{2}Miami University \quad \textsuperscript{3}UCL and Adobe Research
}

\maketitle

\begin{abstract}
The success of StyleGAN has enabled unprecedented semantic editing capabilities, on both synthesized and real images. However, such editing operations are either trained with semantic supervision or described using human guidance. 
In another development, the CLIP architecture has been trained with internet-scale image and text pairings, and has been shown to be useful in several zero-shot learning settings. 
In this work, we investigate how to effectively link the pretrained latent spaces of StyleGAN and CLIP, which in turn allows us to automatically extract semantically labeled edit directions from StyleGAN, finding and naming meaningful edit operations without any additional human guidance. Technically, we propose two novel building blocks; one for finding interesting CLIP directions and one for labeling arbitrary directions in CLIP latent space.
The setup does not assume any pre-determined labels and hence we do not require any additional supervised text/attributes to build the editing framework. 
We evaluate the effectiveness of the proposed method and demonstrate that extraction of disentangled labeled StyleGAN edit directions is indeed possible, and reveals interesting and non-trivial edit directions.
\end{abstract}

\section{Introduction}
\label{sec:intro}
Generative adversarial networks~(GANs) are able to synthesize new images similar to a given set of existing images. After initial research in the fundamental aspects of GAN architecture and training~\cite{radford2015unsupervised, LSGAN2017,WGAN2017, WGANGP2017, gulrajani2017improved}, StyleGAN~\cite{STYLEGAN2018} has emerged as the most popular architecture. Building on pre-trained GANs, recent research proposed several solutions for GAN inversion~\cite{abdal2019image2stylegan, richardson2020encoding, tewari2020pie, zhu2020improved, zhu2020domain} and semantic control~\cite{harkonen2020ganspace,10.1145/3447648,shen2020interfacegan} that together enable semantic editing of existing photographs. One class of algorithms
to provide semantic editing capability is \textit{supervised} using training data to define these semantic concepts, e.g., age, gender, pose, and lighting for face images.
Alternatively, other algorithms~\cite{harkonen2020ganspace} find latent space manipulations in an unsupervised manner but the directions themselves are manually labeled and curated in a post process.
In this work, starting with pretrained GANs, we investigate \textit{if it is possible to both  extract and label disentangled semantic controls without additional supervision.} In particular, we use a linguistic model to to determine which attributes of set of images are important, and we propose meaningful words to describe edit operations.

In another recent advancement, CLIP~\cite{radford2021learning}, embedding spaces have been learned for image/text embedding on very large internet-scale data. The learned spaces, although trained using only loose image-keyword pairing information, have been shown to be rich and effective for several zero-shot tasks, i.e., not requiring any further training or fine-tuning for new tasks. For example, in the context of image manipulation, CLIP and StyleGAN have been utilized for text guided editing~\cite{patashnik2021styleclip} or zero-shot domain transfer~\cite{gal2021stylegannada, zhu2021mind}.

\begin{figure}
    \centering
    \includegraphics[width=\linewidth]{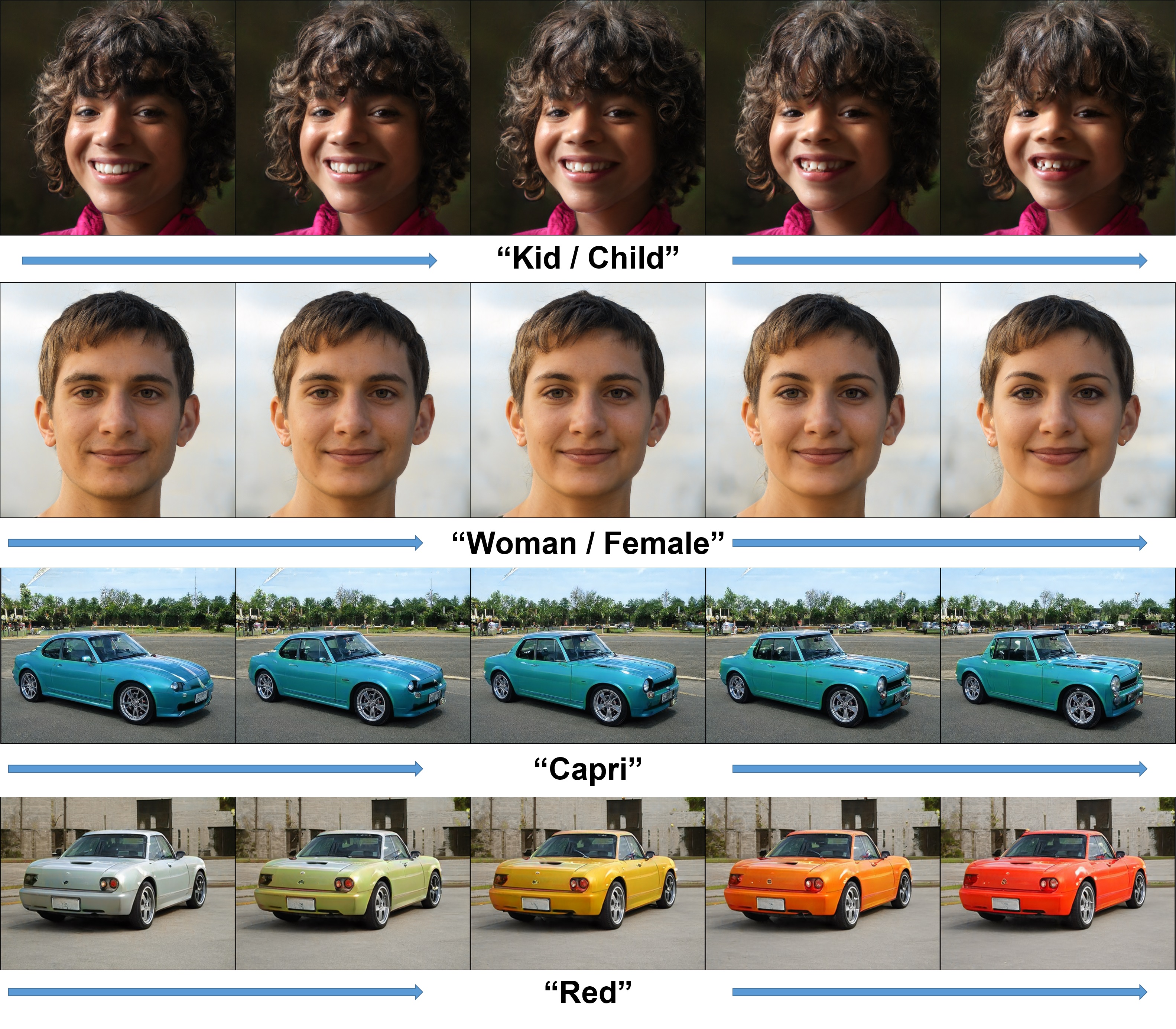}
    \caption{ We propose CLIP2StyleGAN, an unsupervised framework to extract and label disentangled directions in StyleGAN. The figure shows some fine-grained edits and corresponding labels extracted by CLIP2StyleGAN framework for faces and cars.}
    \label{fig:teaser}
\end{figure}

Given the success of the above approaches, we investigate whether the salient attributes of an image dataset can be automatically identified based on CLIP embeddings. Specifically, we explore the following: 
(i)~how can we identify interesting directions, in a data-driven sense, by analyzing the CLIP image space; 
(ii)~if the above directions are entangled according to textual concepts, i.e., according to the CLIP text space, then how to disentangle the directions; and finally, 
(iii)~how to transfer the disentangled directions, along with their concept labels, to the StyleGAN space. 
By developing methods to answer the above questions, we enable unsupervised extraction of labeled StyleGAN edit directions, a process we term as \name to highlight the joint analysis of the CLIP and StyleGAN spaces.

We evaluate \name on two datasets of portraits and cars, we evaluate the quality of the extracted edit directions,  and we provide a user study to compare the words we automatically choose to describe our edits  to human labels. Note that, as a byproduct, our method provides unsupervised classifiers for concepts that are found as disentangled directions. Language is correlated with our ability to perceive attributes such as gender, age, or color~\cite{maier2018native}, so by extracting attributes in a joint text/image space we automatically suggest attributes that are perceptually important for editing, data augmentation, and evaluation of image processing models. Fig.~\ref{fig:teaser} shows top few extracted and labeled directions for the face and the car StyleGAN spaces.

\section{Related Work}

\paragraph{High-Quality GANs.}
Two state of the art architectures for GANs are BigGAN~\cite{brock2018large} for ImageNet and StyleGAN for specific classes, such as faces, cars, and human bodies. StyleGAN was developed over a series of papers by Karras and various co-authors~\cite{karras2017progressive,STYLEGAN2018,Karras2019stylegan2,Karras2020ada,karras2021aliasfree}. The quality of the StyleGAN output is particularly impressive for well curated high-resolution datasets, most importantly FFHQ~\cite{STYLEGAN2018} which focuse on human faces, but also AFHQ~\cite{choi2020starganv2} focusing on animals, 
and LSUN objects~\cite{yu15lsun} which includes images of cars among others. On all of these themed-datasets, StyleGAN yields very good results. In this work, we do our analysis on the FFHQ~\cite{STYLEGAN2018} dataset trained on StyleGAN2~\cite{Karras2019stylegan2}
because it is considered state of the art for face image generation and because as humans we are especially tuned to understand and describe images of faces. Note that StyleGAN3~\cite{karras2021aliasfree} is not published at the time of this submission and the code was released too close to the deadline to switch our implementation.

\paragraph{Joint Visual-Linguistic Models.}
Computer vision and NLP researchers have been interested in generating diverse representations of images by combining natural language and image representations, for example DeVISE~\cite{frome2013devise} was an early work that showed how language could improve a model. Such methods combine visual and language models to perform many interesting downstream tasks such as visual question answering and image captioning. ICMLM~\cite{sariyildiz2020learning} learns visual representations over image-caption pairs to inject global and localized semantic information into visual representations. ViLBERT~\cite{lu2019vilbert} extends the BERT~\cite{devlin2019bert} architecture for learning task-agnostic joint representations of image content and natural language. VL-BERT~\cite{su2020vlbert} also learns a joint representation that is effective for many visual-linguistic downstream tasks. Other notable works in this domain~\cite{li2020oscar,desai2021virtex,tan2019lxmert,chen2020uniter} also perform joint image-text embedding for multi-modal visual and textual understanding. A very recent development in this domain is OpenAI's CLIP~\cite{radford2021learning} model. With its powerful image and text representations, it is able to perform zero-shot transfer in many downstream tasks. We use the CLIP model to analyze the latent space of StyleGAN due to its zero-shot classification/regression and labeling properties which we explore in the Sec.~\ref{sec:method}.

\paragraph{GAN-Based Semantic Editing and Layer Interpretation.}
Understanding and manipulating the latent spaces of GANs has been a topic of recent research interest. Specifically in the StyleGAN domain, various works~\cite{bau2018gan,bau2019seeing,harkonen2020ganspace,shen2020interfacegan,tewari2020stylerig,wang2021sketch} explore latent spaces to enable high quality image editing applications. 
To enable image editing, image embedding is used as a technique to project real images into the GAN's latent space. 
The latent-codes of embedded images have the convenient property that simple manipulations of the latent codes seem to enable the user to edit the real images using the semantic properties of GANs. 
Related to StyleGAN,  Image2StyleGAN~\cite{abdal2019image2stylegan} embeds the images into an extended $W^{+}$ latent-space to which is  capable of representing a diverse set of images with more variety and detail than the original GAN. Other works~\cite{zhu2020domain,richardson2020encoding,tewari2020pie, zhu2020improved, alaluf2021restyle} in this domain use regularizers and encoders built on top of the StyleGAN latent space to maintain the semantic meaning of the embedding by finding the representations close to the original space of StyleGAN. They often include loss terms designed to preserve specific attributes that are meaningful to a particular dataset (e.g. identity for faces) when finding a latent-code. 

Image editing frameworks in the StyleGAN domain~\cite{harkonen2020ganspace,shen2020interfacegan,tewari2020stylerig,10.1145/3447648} analyze the latent space to identify linear and non-linear paths for semantic editing. InterfaceGAN~\cite{shen2020interfacegan}  finds linear directions to edit latent codes in a supervised manner, while GANSpace~\cite{harkonen2020ganspace} extracts unsupervised linear directions for editing using PCA in the $W$ space. Another framework, StyleRig~\cite{tewari2020stylerig}, constructs a riggable 3D model on top of StyleGAN. StyleFlow~\cite{10.1145/3447648}, extracts non-linear paths in the latent space to enable sequential image editing. In the area of text-based image editing, StyleCLIP~\cite{patashnik2021styleclip} uses the CLIP embedding vectors to adjust the latent codes of a GAN. Another CLIP based framework, StyleGAN-NADA~\cite{gal2021stylegannada}, uses the CLIP framework for zero-shot domain adaptation. However, the existing CLIP based image editing approach requires a user to provide a textual description of the edit as input. We aim to automatically identify such descriptions. 

Investigations into the layer representations of GANs have broadened the application of GANs beyond image editing. In the domain of GAN layer interpretation, more recent works have shown that GANs can be used to perform various downstream tasks including few-shot and unsupervised image segmentation~\cite{zhang2021datasetgan,tritrong2021repurposing,Abdal_2021_ICCV, collins2020editing, bielski2019emergence} , extraction of 3D models~\cite{pan2021do,chan2020pigan,meng2021gnerf} and other novel local image edits such as hairstyle manipulation~\cite{zhu2021barbershop, tan2020michigan }.

\begin{figure*}[t]
    \centering
    \includegraphics[width=\linewidth]{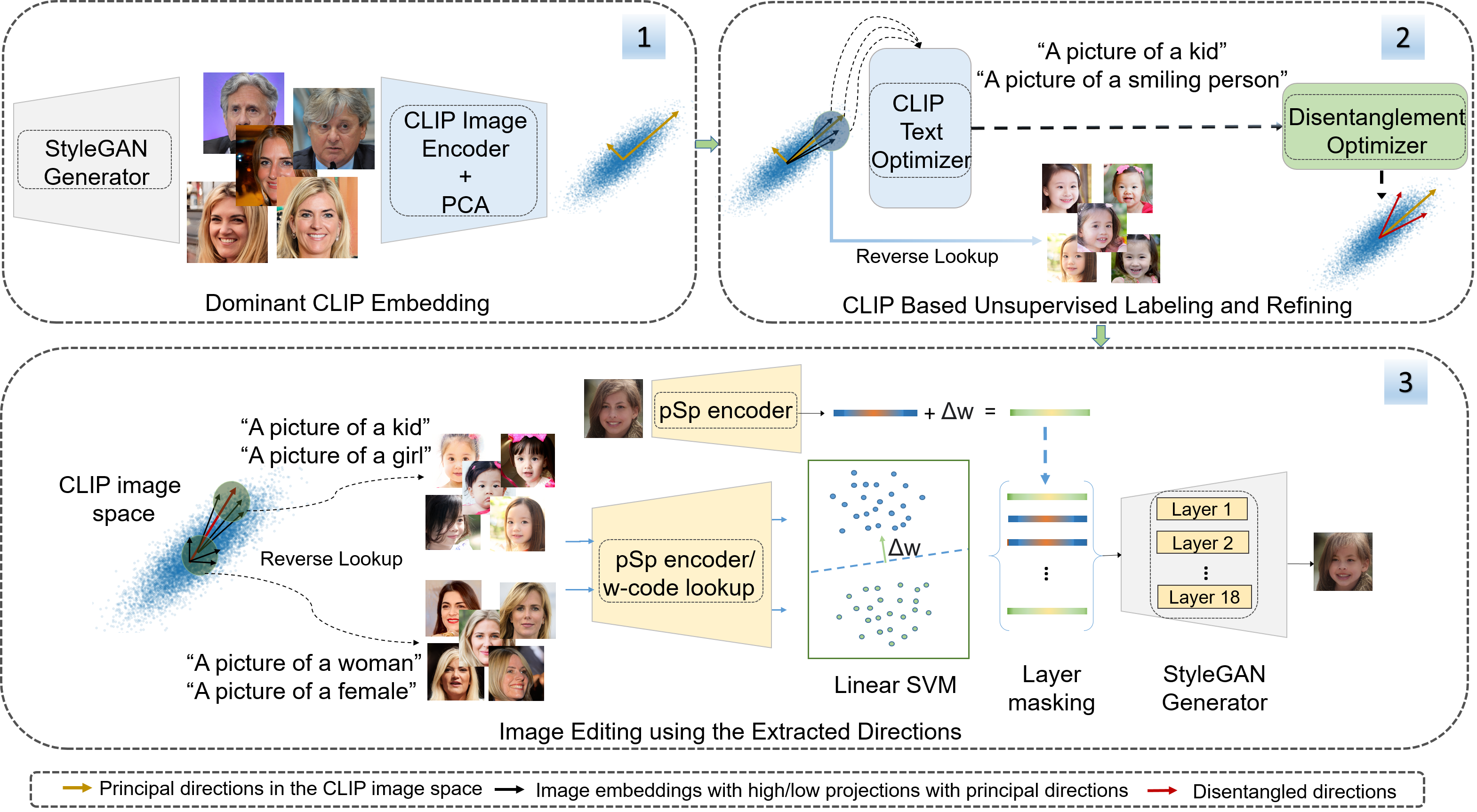}
    \caption{Overview of \name pipeline. We produce disentangled and semantically-labeled image edit directions via an unsupervised joint analysis of the CLIP latent spaces, both image and text, and the StyleGAN latent space. Empirically, we found the extracted directions to be universal and can directly be used to edit real images (see Fig.~\protect\ref{fig:teaser}).  }
    \label{fig:pipeline}
\end{figure*} 

\section{Method}
\label{sec:method}
Our proposed method for unsupervised discovery and labeling of StyleGAN edit-directions is divided into three steps. First, using the CLIP image space, we compute semantic directions based on the analysis of a set of images, e.g., from a dataset or real images like FFHQ (70k images) or synthetic StyleGAN-generated images (arbitrarily many, e.g. 100K). Second, we disentangle the candidate directions and label them using the CLIP text encoder. Third, we map the labeled disentangled directions to the StyleGAN latent space to perform various unsupervised semantic edits.  Fig.~\ref{fig:pipeline} shows the pipeline of our framework.

\begin{figure}[t!]
    \centering
    \includegraphics[width=\linewidth]{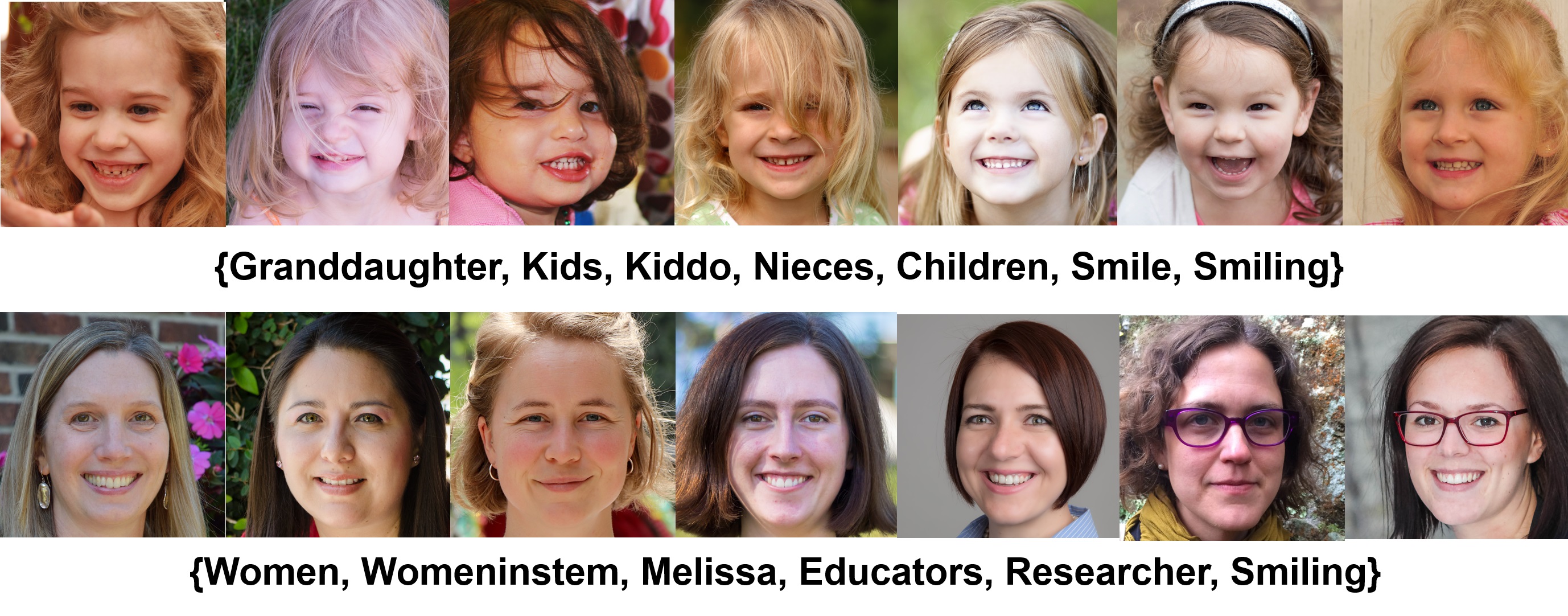}
    \caption{The first and second principal direction extracted from the PCA analysis of CLIP image embeddings of the FFHQ dataset. }
    \label{fig:ffhq_proj}
\end{figure}

\subsection{CLIP Based Direction Extraction}

Given a set of $n$ different images, the goal of this stage is to compute a set of directions in the CLIP latent space that capture statistically significant variations within the dataset. 
The set of images is embedded into CLIP latent space using the pre-trained image encoder with the ViT-B/32 implementation, which embeds the images (or alternatively text) into $512$ dimensional space.

We first describe multiple options to compute such directions, and then provide analysis specific to our application.

\paragraph{Dominant CLIP Embedding.}
The CLIP image encoder~\cite{radford2021learning} is trained on the common-crawl dataset, an internet-scale set of images that encompasses a broad range of visual concepts. However, a typical high-quality GAN would be trained on a more specific set of images, for example human faces in FFHQ~\cite{STYLEGAN2018}, or vehicles in LSUN-Cars~\cite{yu15lsun}.  Images in such datasets are expected to share a common directional component in CLIP space, a vector pointed at the center of the CLIP embeddings of images in the dataset. The theme of these datasets exhibits itself in the mean vector, $\vb*{\mu}$, of their CLIP embeddings.  For the FFHQ dataset, the vector $\vb*{\mu}$ is most similar to CLIP embedded tokens of \textit{a picture of a person, individual, adult,} or \textit{headshot}.  Hence, these are not useful as edit directions, so we subtract the mean from the CLIP vectors. Hence, for each (candidate) edit direction $\vu{u}$, the CLIP embeddings associated with that direction are the latent codes of the ray $\vb*{\mu} + \alpha \vu{u}$ with $\alpha \in \mathbf{R}$.

\paragraph{Direction Computation.}\label{sec:direction-computation}
We consider three different options to extract CLIP latent space directions from a set of images: (i)~random, (ii)~PCA, and (iii)~ICA, or hybrid sets that combined directions. The common assumption of these algorithms is that the CLIP latent space is a semantic directional space, and we expect to see an evolution of attributes along the extracted directions, \eg, age progression from a young person to an old person. We discuss the random and ICA approaches in the supplemental, and focus here on PCA.

\paragraph{PCA.} Principal Component Analysis~(PCA) assumes the underlying data comes from an approximately Gaussian distribution. The output of the PCA is a set of 512 orthogonal principle component axes, which we will use as editing directions. In addition, each direction is associated with a variance that can be used to sort the directions.

\paragraph{PCA + Random Directions (Hybrid).}
In principle, our approach works for any set of candidate vectors. 

For example, one can use the initial 10 PCA axes, and then pick random vectors from the remaining subspace as additional directions. We do this by picking examples that have low correlation with the initial PCA axes (\eg, Beard + Glasses in Sec.~\ref{sec:results}).

\subsection{CLIP Based Unsupervised Labeling}

 Let the set $\mathcal{X} := \{\vb{x}\in \mathbb{R}^{512}\}$ indicate a set of $n$ different CLIP embeddings of images either from a real dataset (e.g., FFHQ) or by random sampling of StyleGAN.  In our experiment, we used $n=70$k in case of FFHQ and  $n=100$k in case of StyleGAN generated images . We use the notation $\vu{x}$ to indicate a normalized vector $\vu{x} := \vb{x}/\|\vb{x}\|$. We also use a set of directions $\mathcal{U} := \{\vu{u} \in \mathbb{R}^{512}\}$ indicating potential semantic-attributes predicted using one of the methods described in Sec.~\ref{sec:direction-computation}.  
 In the following, given an arbitrary candidate edit-direction $\vu{u} \in \mathcal{U}$, our goal is to find a set of words, from a given lexicon, that describe the edit direction. In addition, we find subsets of $\mathcal{X}$ that are positive examples $\mathcal{X}^+$ and negative examples $\mathcal{X}^{-}$ of the attribute.  

As described earlier, we use a translational component, $-\vb*{\mu}$, to subtract away any common attributes of the dataset as a whole.

We identify samples in $\mathcal{X}$ that are not relevant for a given attribute direction.
Specifically, we ignore any vectors $\vb{x}\in\mathcal{X}$ as irrelevant if, after translation by $-\vb*{\mu}$, they are not positively correlated with the direction $\vu{u}$. 
Then, we leverage the CLIP text encoder to label the directions.
First, all the relevant vectors are sorted based on the projection $\vu{u}\vdot(\vb{x}-\vb*{\mu})$ and then we sample 100 top and bottom CLIP image embeddings based on these projections. These two groups of CLIP embeddings form the positive and negative examples of a given attribute, $\mathcal{X}^{+}$ and $\mathcal{X}^{-}$, and they are the basis of a potential edit.

Rather than naively use $\vu{u}$ as the edit direction, we instead use a modified vector that more directly aligns with the vectors in $\mathcal{X}^+$.

Motivated by zero-shot classification approaches\cite{radford2021learning}, we find the centroid of the normalized values $\vu{x}$ in the CLIP image space, and project the centroid onto a hypersphere as, 
\begin{align}
    \mathbf{x}_{m}^+ &= \expval{\vu{x}}/\|\expval{\vu{x}}\|, \quad \vb{x} \in \mathcal{X}^+
\end{align}
where $\expval{\cdot}$ is the expected value operator, yielding target of our edit as the direction $\vb{x}_m^+$.

\begin{algorithm}[b!]
\SetAlgoLined
 \KwIn{A CLIP vector $\mathbf{x} \in \mathcal{R}^{512}$;}
 \KwIn{Token embeddings $\vb{E}\in\mathcal{R}^{m\times512}$ }
 \KwIn{The number of GD steps (\texttt{maxit}) }
 \KwOut{the predicted set of labels $\mathcal{L}$  }

\tcc{Gradient Descent}
 Initialize $\vb{z} \in \mathcal{R}^m \leftarrow \va{0}$\;

\For{$i \in 1 \dots \texttt{maxit}$}{ 

   $\mathbf{z} \leftarrow \mathbf{z} - \eta \nabla_\mathbf{z} L(\vb{z}, \vb{x}_m)$\;
}

$\vb{e} \gets \vb{E}^T\sigma(\vb{z})$\;
$s_i =   \vb{e}_i^T \vb{e},
\quad \vb{e}_i \in \text{rows}(\vb{E})$\; 
$\{i_1,\dots,i_k\} \leftarrow \textsc{TopK-Indices}(s_i)$\;
$\mathcal{L} \leftarrow \{\vb{e}_{i_1}, \dots \vb{e}_{i_k}\}$\;

 \caption{Text prediction using CLIP text encoder}
 \label{alg:text}
\end{algorithm}

Next, we define an optimization  that uses the CLIP text encoder and a lexicon of $m$ different potential labels to automatically generate textual descriptions of the directions. 
 
Our aim is different from captioning or explaining images, we are only interested in finding individual tokens as words to use as a label for an edit direction in clip space. 
The input to text encoder $T_E$ is a vector sequence from token-space, including a prefix such as `a picture of a' or `a picture of a person with a' followed by the embedding of a word-token ($\vb{e}$). 
Then the CLIP-embedding of the token is 
\begin{align}
  \vb{t} &= T_E(\text{prefix} \oplus \vb{e}),  \label{eq:t}
\end{align} 
where $\oplus$ concatenates the sequence of prefix tokens and ends with a token  $\vb{e}$ from our lexicon.  
Our goal is to find a word token $\vb{e}$ so that its CLIP embedding $\vb{t}$ is aligned with an edit direction in CLIP space ($\vb{x}_m^+$).
However, the token vector $\vb{e}$ which best matches will, in general, not be in our lexicon. 
To address this issue, we define a soft selection variable vector ($\vb{z} \in \mathbb{R}^{m}$) 
such that,  
\begin{equation}
\vb{e} = \vb{E}^T\sigma(\vb{z}),
\end{equation}
where $\sigma$ is the sigmoid function and $\vb{E}$ is the embedding layer of the CLIP text encoder, a matrix with $m$ rows and 512 columns .

Our goal is to find as-sparse-as-possible of a selection vector $\vb{z}$. 
We consider the entropy of the selection vector $\vb{z}$ to be an indication of the complexity of the edit described by $\vb{e}$, and so regularization by the entropy will bias the corresponding edit vector $\vb{t}$ in CLIP space towards simpler concepts. 
This loss function ($L$) is used to minimize the cosine distance between $\vb{t}$ and $\vb{x}_m$ while also keeping the entropy of $\vb{z}$ low. 
The objective function can be written as :
\begin{align}  
     \mathbf{z} &= \argmin_{\mathbf{z}_k} L(\vb{z}, \mathbf{x_{m}}) 
    \label{eq:loss_func} \\
     &= \argmin_{\mathbf{z}_k} \left( d_{\cos{}}(\vb{t}, \vb{x_m}) + \lambda H(\sigma(\vb{z})) \right) \nonumber
\end{align}
where $\vb{z}$ is the soft-selection vector, $\vb{t}$ is defined by equation~\eqref{eq:t}, and $H$ is the entropy. Although similar, the vector $\vb{t} \neq \vb{x}_m$ due to regularization, and we use $\vb{t}$ as an improved edit direction because it is a mixture of a small number of tokens in the lexicon.

Once a suitable token ($\vb{e}$) and the corresponding clip vector ($\vb{t}$) have been chosen, we identify a set of candidate words $\mathcal{L}=\{\vb{e}_{i_1} \dots \vb{e}_{i_k}\}$ where  $i_1\dots i_k$ are the indices of the top-$k$ tokens in our lexicon in descending order of their inner product with $\vb{e}$. Note that we use an inner product and not the cosine similarity so that we give preference to word embeddings with a larger magnitude, which often corresponds to the specificity and importance of a word~\cite{schakel2015measuring}. 

In addition, several prefixes can be considered in equation~\ref{eq:t}. In this case, we repeat our entire process for each prefix, and then set $\mathcal{L}$ to the union over all prefix prompts. Our approach for extracting labels is shown in Algorithm~\ref{alg:text}.

\begin{figure*}[h]
    \centering
    \includegraphics[width=\linewidth]{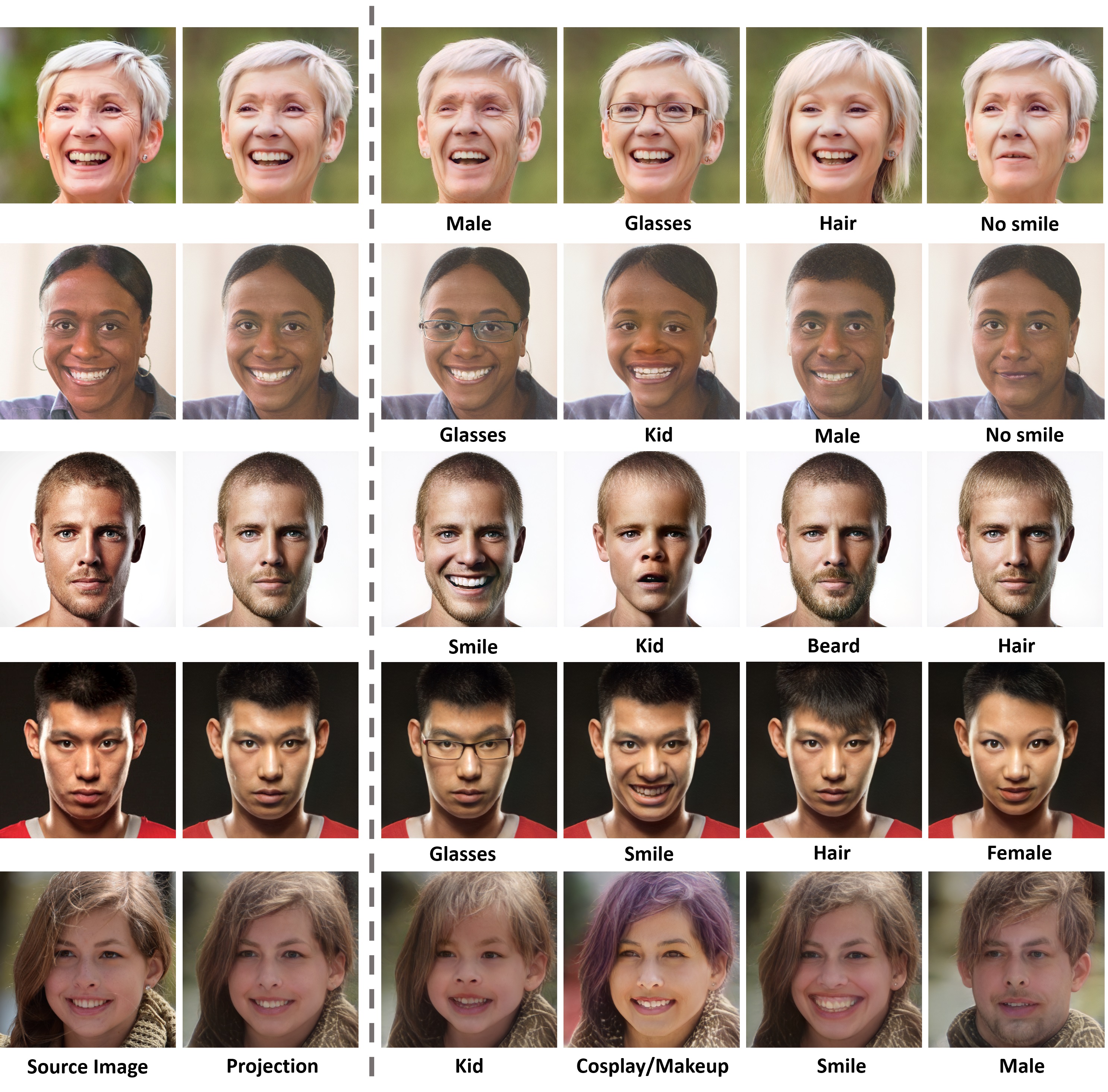}
    \caption{Editing real images using the labeled directions produced by \name. Additionally, we evaluate the correlation of the edits with extracted text attributes via a user study. All edits are applied to the corresponding projected images. 
    }
    \label{fig:edits}
\end{figure*}

\subsection{Refining Labels}
\label{sec:refine}

The CLIP directions, by themselves, can represent multiple attribute changes, \eg, change in age and expressions predicted by Algorithm~\ref{alg:text}.
The edit-direction we automatically identify in this case may combine them into a single vector, and the list $\mathcal{L}$  will contain tokens describing each attribute. 
To encourage disentanglement, we instead represent that direction as the sum of two or more atomic directions. 
We do this by clustering words in $\mathcal{L}$ that have similar meanings, and then replacing the original edit direction with a set of edit directions aligned to each subset of words. 
Specifically, we use a Wu-Palmer word similarity score~\cite{wu1994verb} (maximized over all potential senses of a word) to determine if textual labels are related (e.g., synonyms or child-concepts) or not.  Then, greedily we select a word from the list and then remove all similar words with (matching) scores greater than $0.9$. We repeat the process for the next word in the modified list, until we exhaust the list. 
If more than one word remains in the list, then we consider the edit-direction to be entangled because it combines multiple linguistic concepts.  Our aim is to replace it by multiple (atomic) vectors that are each better aligned to the CLIP embeddings of the individual words in $\mathcal{L}$. 
We consider two approaches for generating edits based on the new tokens.

In the first approach, we simply abandon the direction $\vu{u}$ by removing it from the set $\mathcal{U}$, and then adding new directions based on the words in $\mathcal{L}$.

Then the process can simply resume with the additional edit directions  which represent more refined and disentangled concepts.

In the second approach, we use gradient descent to find new vectors  corresponding to semantically different words in $\mathcal{L}.$ The objective function for this optimization is represented by:
\begin{align}  
\mathbf{B} 
        &=  \argmin_{\mathbf{B}} L_\text{split}(\vb{B}, \vu{u}, \vb{w}, \vb{T})  \label{eq:disentangled}\\
        &= \argmin_{\mathbf{B}} \beta L_\text{rec}(\vb{B},\vu{u},\mathbf{w})  
           + L_\text{indep}(\vb{B})
           + L_\text{tok}(\vb{B}, \vb{T}),   \nonumber
\end{align}
where $\mathbf{B}$ is a matrix representing the disentangled vectors along columns, $\hat{\mathbf{u}}$ represents the given principal direction, $\mathbf{w}$ represents a row vector of confidence scores predicted by the CLIP text encoder for the attributes to be disentangled, and $\vb{T}$ is a matrix whose columns are the CLIP embedding of the reduced set of words in $\mathcal{L}$. The loss function combines a reconstruction term 
\begin{align}
     L_\text{rec}(\vb{B}, \vu{u},\vb{w})  &:= \| \hat{\mathbf{u}} - \mathbf{B}\mathbf{w}  \|
\end{align}
which is minimized when the original vector $\vu{u}$ is in the column space of $\vb{B}$; an independence term 
\begin{align}
L_\text{indep}(\vb{B})   &:=  \| \mathbf{B}^T  \mathbf{B} - \mathbf{I} \|_F 
\end{align}
which is minimized then the columns of $\vb{B}$ are independent; and a token-alignment term
\begin{align}
     L_\text{tok}(\vb{B},\vb{T})  & := -\Tr( \mathbf{B}^T\mathbf{T} ), 
\end{align}
which is minimized when the new directions are parallel to the vectors $\vb{t}_i$ corresponding to the different words in $\mathcal{L}$. 
\subsection{Image Editing using the Extracted Directions}
\label{sec:project}
Once an extracted CLIP direction has been disentangled and labeled, our approach to image editing is very similar to the \textit{global} style edit  described StyleCLIP~\cite{patashnik2021styleclip}.
In order to project CLIP extracted labeled directions, we simply embed the corresponding negative and positive example images in $\mathcal{X}^+$ and $\mathcal{X}^-$ in the StyleGAN $W^{+}$ space using the pSp~\cite{richardson2020encoding} encoder, if the direction is extracted from the FFHQ dataset. In case of the StyleGAN generated images, we simply use the $W$ codes of the corresponding images. 
Then, we compute a unit direction in the StyleGAN space using linear SVM on the embedded latents, as was done in InterFaceGAN~\cite{shen2020interfacegan}, in order to determine an edit direction.

\begin{figure}[h]
    \centering
    \includegraphics[width=\linewidth]{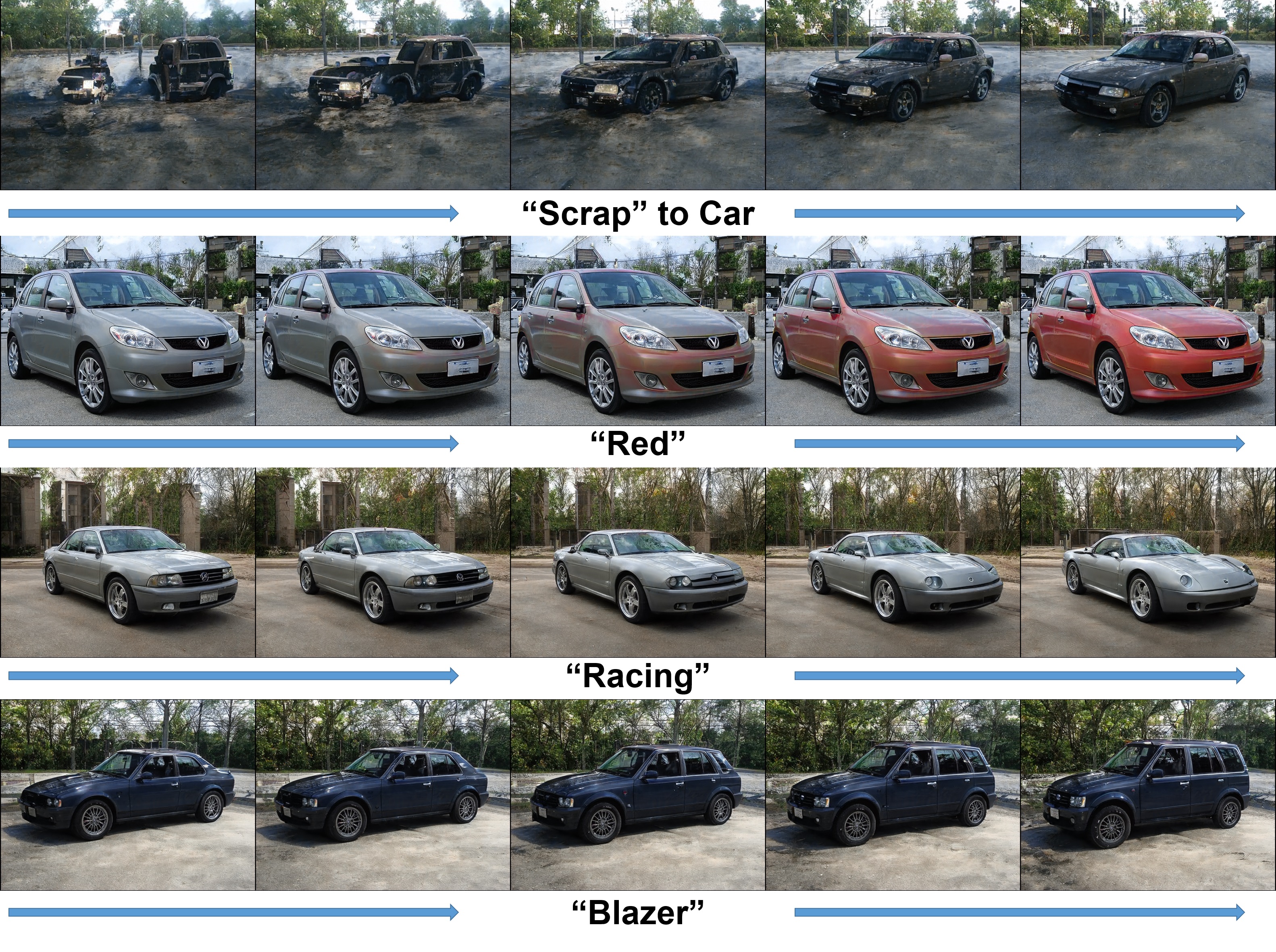}
    \caption{Edits performed on cars using the labeled directions extracted by \name.}
    \label{fig:car_edits}
\end{figure}

\section{Results}
\label{sec:results}

\begin{table}[t!]
\caption{ Evaluation of the labeled directions. The left column list sets positive examples of an edit using our target $\mathcal{X}^+$ set and two different methods. The remaining columns are the score of a zero-shot classifier that predicts which of the three labels is correct. The second column is our automatically chosen word. The third column is from a manually chosen keyword, and the last column is a manually chosen negative keyword. The best score among GANSpace vs. ours is shown in bold.
} 
\centering
\scriptsize
\begin{center}
    
\def\ua{\ensuremath{\uparrow}}
\def\da{\ensuremath{\downarrow}}

\begin{tabular}{rccc} \toprule

 \multicolumn{1}{c}{Image-Sets}   & 
  \multicolumn{3}{c}{CLIP ZS Classifier Score}
 \\ 

  & Kids\ua & Teens\ua & Adults\da 	 \\  
CLIP($\mathcal{X}^+$)  &0.8462 &0.1257 &0.0281  	 \\  
SG-ours  &\textbf{0.8252}	&0.1266  &\textbf{0.0480} 		 \\  
GANSpace  &0.3442	&\textbf{0.4932}  &0.1626 		 \\  
\midrule
  &Beard\ua &\ Facial-Hair\ua & Clean-Shaved\da	 \\
CLIP($\mathcal{X}^+$)  &0.8052 &0.1913 &0.0033   \\
SG-ours  &0.5486	&\textbf{0.4465} &\textbf{0.0146}\\
GANSpace  &\textbf{0.6328}	&0.3284 &0.0386	 \\
\midrule

  &Smile\ua & Happy\ua &Sad\da \\
CLIP($\mathcal{X}^+$)  &\textit{0.9170}	&\textit{0.0814}  &\textit{0.0019}  	 \\
SG-ours  &\textbf{0.9092}	&\textbf{0.0859}  &\textbf{0.0047} 	 \\
GANSpace  &0.9082	&0.0858  &0.0059	 \\
\midrule
 &Male\ua & Masculine\ua  &Female\da  \\
CLIP($\mathcal{X}^+$)  &0.7139	&0.2751 &0.0108  	 \\
SG-ours  &\textbf{0.6914}	&0.2837 &\textbf{0.0248} \\
GANSpace  &0.5811	&\textbf{0.3870}  &0.0323\\

\bottomrule
\end{tabular}
\end{center}
\label{tab: text_preds}
\end{table}

\begin{table}[h]
\caption{Classification scores before and after the disentanglement of edits, evaluating how well the directions get disentangled. A + B: Entangled direction;  A: First disentangled direction; B: Second disentangled direction; $B + G$: Beard + Glasses; $ K + S$: Kids + Smile; $S_1$, $S_2$: CLIP zero-shot scores predicting which of the two words match the images. } 
\centering
\scriptsize
\begin{tabular}{rrr|rrr|rrrrr} \toprule

\multicolumn{1}{c}{A + B}   & \multicolumn{1}{c}{$S_1$}  & \multicolumn{1}{c}{$S_2$} & 
\multicolumn{1}{c}{A}   & \multicolumn{1}{c}{$S_1$}  & \multicolumn{1}{c}{$S_2$} & 
\multicolumn{1}{c}{B}   & \multicolumn{1}{c}{$S_1$}  & \multicolumn{1}{c}{$S_2$} 
\\ \midrule
$B + G$ &$0.30$ &$0.70$ & $B$   &$1.00$	&$3e^{-5}$  &  $G$ &$2e^{-4}$  &$1.00$	 	 \\
\midrule
$K + S$ &$0.33$ &$0.66$ & $K$ &$0.99$ &$1e^{-3}$ & $S$ &$1e^{-3}$ &$0.99$ 
\\
\bottomrule
\end{tabular}
\label{tab: table_disentangle}
\end{table}

\begin{table*}[htb!]
    \centering
    \caption{Top-5 User-generated words for the predicted edit directions sorted by frequency, which is shown in parenthesis.  Relevant words are indicated in bold. The total  frequency of words that match our labels is shown parenthesis in the left.  }
    \begin{tabular}{r|lllll}
         Our Words     &  \multicolumn{5}{c}{Top-5 User Generated Words} \\ \hline
         youth,children \textbf{(69\%)}    &  \textbf{child(45\%)}  &  mouth(25\%)&    \textbf{age(24\%)}&     glasses( 1\%)&   old( 1\%) \\
         frames,glasses \textbf{(75\%)}    &  \textbf{glasses(75\%)}&  hair(10\%)&     look( 9\%)&      lips( 2\%)&      eyes( 1\%)\\
         female,women \textbf{(63\%)}      &  \textbf{gender(34\%)} &  \textbf{female(15\%)}&  hair(14\%)&     \textbf{woman(14\%)}&    mouth( 5\%) \\ 
         haired,bearded \textbf{(87\%)}    &  \textbf{beard(85\%)}  &  older( 6\%)&    mouth( 2\%)&    \textbf{hairier( 2\%)}&  masculine( 1\%) \\  
         male,adult \textbf{(78\%)}        &  \textbf{gender(38\%)} &  \textbf{man(24\%)}&     hair(18\%)&     \textbf{male(16\%)}&     teeth( 4\%) \\
         laugh,teeth,smile \textbf{(88\%)} &  \textbf{smile(66\%)}  &  \textbf{teeth(12\%)}&   \textbf{happier(10\%)}&  eyes( 3\%)&     woman( 1\%) \\ \hline
    \end{tabular}
    \label{tab:user-study-ranked-words}
\end{table*}


\subsection{Implementation Details}
We conduct our experiments on an A100 GPU. For the text optimization algorithm we use the ADAM optimizer with an initial learning rate of $5e^{-3}$. We perform optimization for $150$ steps with $\lambda$ set to 1. 
Alternatively, an $L1$ regularizer can also used on $\sigma(\vb{z})$ for which $\lambda$ is set to $1e^{-4}$. 
The algorithm takes under $1$ minute to converge. For the additional disentanglement step in Eq.~\ref{eq:disentangled}, we selected $\beta$ to be $0.1$ and we use the ADAM optimizer with initial learning rate of $1e^{-3}$.

\subsection{Qualitative Results}

We visualize projections of the data (FFHQ images) based on the method described in Sec.~\ref{sec:method} given a direction in the CLIP space. 

Fig.~\ref{fig:ffhq_proj} shows some examples of the extracted directions and predicted labels using the first two principal directions of the PCA analysis. We then use the method described in Sec.~\ref{sec:refine} to disentangle the directions. For example, we derive two directions from the first principal direction encoding ``Kids" and ``Smile" which are then projected to the StyleGAN $W^{+}$ space using the method in Sec.~\ref{sec:project}.

Fig.~\ref{fig:edits} shows the results of disentangled edits performed on real images projected in the $W^{+}$ space of StyleGAN. 

 Note that the edits in the Fig.~\ref{fig:edits} show that the algorithm successfully transfers high quality edits from CLIP space to the StyleGAN space.
In order to validate our method on another dataset, we repeat the same analysis using a StyleGAN2 model trained on the LSUN-CAR dataset to extract directions and labels. Fig.~\ref{fig:car_edits} shows some edits in the StyleGAN space using our hybrid directions. The results show some surprising edits of reconstructing cars from scrap. See supplementary material for more details about the extracted directions.

\subsection{Quantitative Results}
We perform four types of quantitative evaluation of our method. First, we evaluate the quality of the edits in StyleGAN space. Second, we evaluate the disentanglement step of our method. Third, we perform the user study to evaluate the performance of our labeling algorithm. We evaluate the performance of the identity preservation of our edits in supplementary materials.

\subsubsection{Editing Quality}
To quantitatively evaluate the quality and accuracy of edits and transfer of directions from CLIP image space to the $W^{+}$ space, we perform some evaluations per edit on the pairs of original and edited real images in the $W^{+}$ latent space. We identify GANSpace~\cite{harkonen2020ganspace} as the closest (unsupervised) method to compare with. Note that in GANSpace the labels are assigned manually and all directions are curated.

First, in order to evaluate if the directions are successfully projected from the CLIP image space to the StyleGAN $W^{+}$ space, we use the CLIP text encoder scores to classify images that are the result of applying an edit in $W^+$ space. 
The CLIP~\cite{radford2021learning} method describes a Zero-Shot prediction method that can be used to assign a score to any set of image embeddings given a textual prompt. Higher scores mean that the images match the prompt. We compare these scores for (i) the word we automatically identified, (ii) a manually chosen word for the edited attribute, and (iii) a word that does not describe the edit, which we expect to have low scores.
For example, in case of `Kids' direction, we compute the scores for `Kids', `Teens' and `Adult'.   
We use the Zero-Shot scores to compare three sets of images; (a) the positive images in $\mathcal{X}^+$ as a baseline, (b) the results of applying an edit using our approach in StyleGAN $W^+$ space, and (c) the result of a GANSpace edit that was labeled with the same attribute by~\cite{harkonen2020ganspace}. The results are shown in Table~\ref{tab: text_preds}.
Notice that the scores of the CLIP vectors in $\mathcal{X}^+$ and our edits are similar to each other. This shows that the direction and the corresponding label was successfully transferred from the CLIP space to the StyleGAN's $W^{+}$ space.

Second, in order to determine if the output directions from the refining step Sec.~\ref{sec:refine} lead to disentangled directions, we show in Table~\ref{tab: table_disentangle} the CLIP zero-shot classification scores for the `Beard + Glass' and `Kids + Smile' before and after the disentangling step. The results show that the directions are, in fact, disentangled from the original entangled vector.

\subsubsection{User Study}

We compare the quality of our generated labels using a mechanical turk `image tagging' task. We evaluated six different automatically generated edit directions and descriptive tags. Each image was randomly selected from the FFHQ dataset, and 342 image pairs were produced by applying one of the six edits.

Each image pair was presented to three different turkers, and the turkers were asked to provide three English words that describe how the edited image is different from the original.  A total of 1,026 surveys were completed. 
For each task, we collect all the words provided by turkers.

We count the frequency of each keyword given by the turkers for a given edit. Then we merge words whose Wu-Palmer similarity score~\cite{wu1994verb} exceeds a threshold (0.8). Finally, we manually determine which words are consistent with the words (in bold Table~\ref{tab:user-study-ranked-words}). We compare these words with some labels generated by our method  given an edit. Table~\ref{tab:user-study-ranked-words} shows the evaluation of the user study. Here, the sum of the relative frequencies of all consistent words is an indicator of the disentanglement of our direction, and the relevance of our generated label. Note that a good direction has a high score.

\subsection{Limitations}
Our method has two important limitations. (i)~The lexicon of text used in CLIP contains sensitive words. We manually filter words that are offensive. (ii)~The directions may also be labeled with stereotypical or prejudicial terms, e.g., assuming the profession of a person. This is a noteworthy (bias) aspect of CLIP space that our method can help to analyze. For the results in this paper, we also manually filter these offensive labels. We discuss this issue in more detail in our section about broader societal impact in supplementary materials. While our method inherits this limitation, at the same time, our method is a good tool to reveal and analyze dataset bias.

\section{Conclusions}
\name is a framework to extract meaningful editing directions in StyleGAN latent space. We introduce two new technical building blocks: a method to extract important directions in CLIP space and an algorithm to label arbitrary directions. After mapping directions to StyleGAN latent space they can be used for editing exiting photographs.
In future work, we would like to use our tool for an extensive semantic analysis of CLIP latent space.

\section{Acknowledgement}

We would like to thank Visual Computing Center (VCC), KAUST for the support, and gifts from Adobe Research. We would also like to thank OpenAI for the CLIP model.

\appendix
\section{Societal and Broader Impact}
\label{sec: impact}

\begin{figure}[b!]
    \centering
    \includegraphics[width=\linewidth]{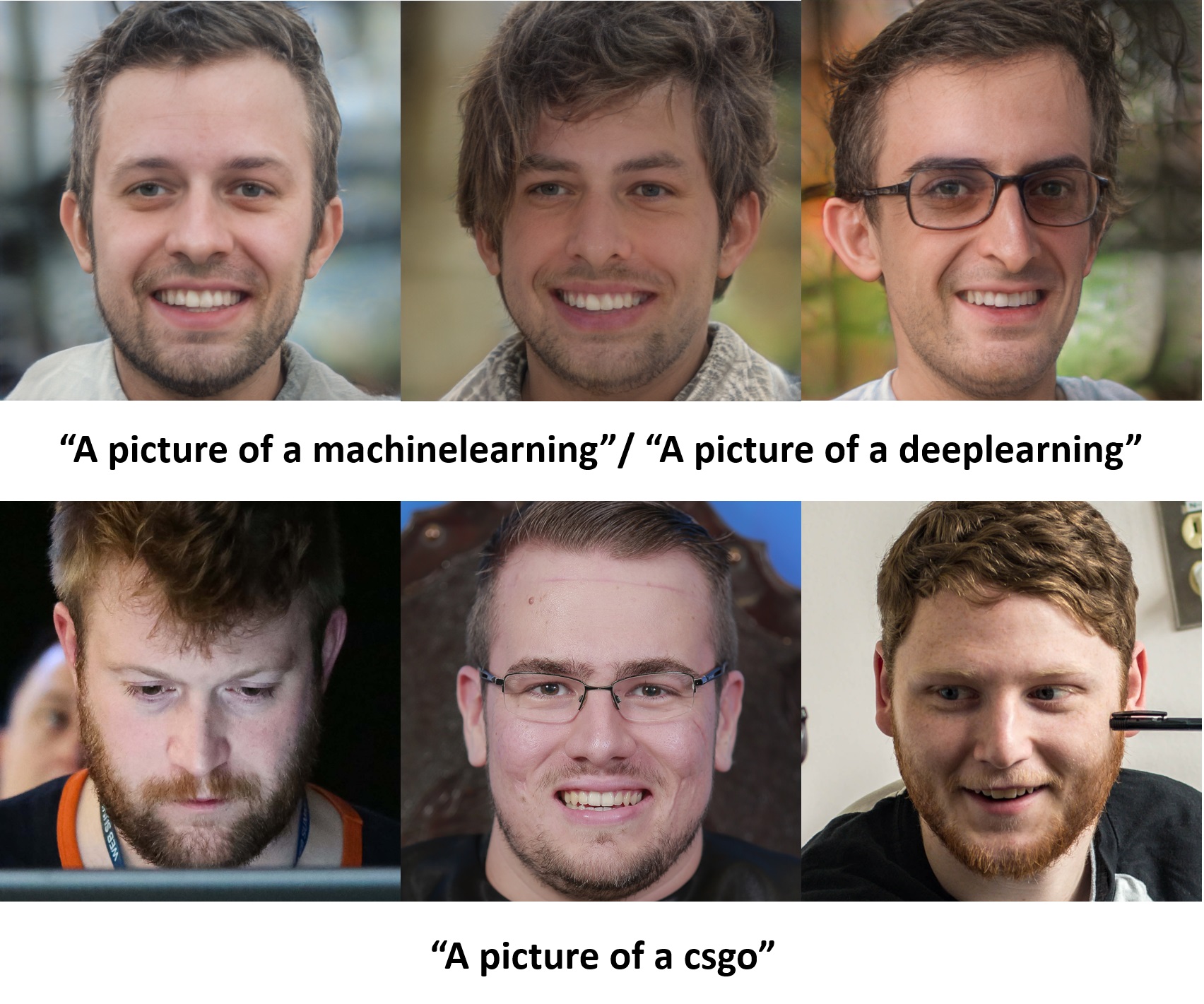}
    \caption{Some examples of data biases present in CLIP training set, where the CLIP zero-shot prediction scores for each of the images have lower confidence score for the prompt, `A picture of a male' than the ones shown in this figure. 
    }
    \label{fig:example}
\end{figure}

Our work builds on the CLIP model and therefore inherits any biases from the dataset used to train the publicly available pre-trained CLIP model.
As a baseline experiment, we used the cosine similarities between the normalized image encoding and text encodings of many prompts. We then looked at the top 100 highest ranked results for each prompt separately. Looking at the results, we made the following observations:
\begin{itemize}
    \item names (i.e., proper nouns) occur very frequently and they dominate the list;
    \item professions like \emph{Engineer} or \emph{YouTuber} and technical terms like \emph{Deep Learning} or \emph{Machine Learning} occur among the highly ranked labels; 
    \item text-based emoticons (e.g., smiley) can also occur; 
    \item results can contain slurs; 
    \item prompt engineering only helps to a minor degree. For example, using a prompt like \emph{`A picture of a man with \{\} hair'} (where \{\} is the query to be replaced) would not restrict highly-scoring text to attributes describing hair. Any sort of token including names, professions, etc.~would still occur as highly ranked.  As an example, in Fig.~\ref{fig:example} we show two group of images. Here for the images in the first row, the CLIP zero-shot prediction scores using the prompt `A picture of a machine-learning'/`A picture of a deep-learning' has a significantly higher confidence score than the prompt,`A picture of a male'. Similarly, for the images in the second row, `A picture of a csgo' gets a significantly higher confidence score than the prompt, `A picture of a male'.
\end{itemize}
Overall, we concluded that without strongly filtering the possible classes (text tokens), a naive zero-shot learning approach using CLIP would be unusable for human portrait images.
The experiment above was applied to individual images, however our approach is applied to the mean of a set of images, which helps to prevent overly specific labels from being discovered.  
Averaging the results over multiple prompts helps a bit, but generally CLIP seems to ignore the grammatical structure of the prompt and the top-hundred list does  change significantly.

\begin{figure}[h]
    \centering
    \includegraphics[width=\linewidth]{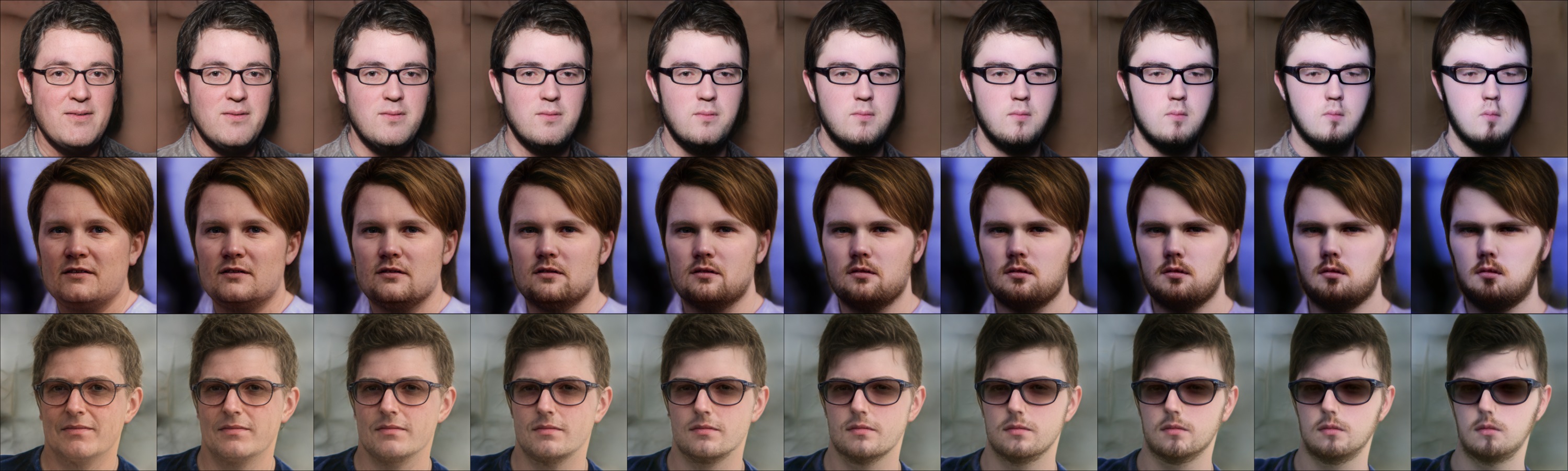}
    \caption{`csgo' edit in the StyleGAN space. 
    }
    \label{fig:csgo}
\end{figure}

Unsuitable labels can be especially controversial when paired with images of humans. Some dominant directions extracted from CLIP by \name are related to different ethnicities and can associate stereotypical/offensive labels to different races and people with different sexual orientations. \textit{In general, any automatic labeling approach that is allowed to draw from an unconstrained set of labels can produce results that are considered offensive, when applied in the context of humans.} Therefore, the labels have to be filtered or be restricted from the original list used in CLIP. However, going in a different direction, not filtering the list of possible labels makes our algorithm well suited to reveal biases in the model and the dataset. This will require careful analysis and thoughtful solutions that we expect to get a lot of scrutiny in future works, from experts not only from the vision and ML communities but involving a much more diverse team. At the moment, we only present results after filtering the labels.

\section{Salience of Prompts}
In CLIP space, words alone do not have a useful embedding, instead one needs a prompt such as `A picture of a \{\}'. However, we find that when arbitrary words are used in the prompt, a meaningless sentence can result. For example, `A picture of a machine-learning', or a `A picture of a smiley'. Also not useful are the names of specific people. We partially address this challenge,  by removing proper nouns and non-English tokens from the lexicon.  Also, we currently have no control over how specific a word is; for example in Fig.~\ref{fig:example} the word `csgo' may make sense in the prompt but it is very specific and may have a high cosine similarity to a generated edit direction. We can also project these directions to the StyleGAN space to visualize the edits. Fig.~\ref{fig:csgo} shows the `csgo' edit in the StyleGAN space. Either due to chance or bias in the density of CLIP space, these specific terms can be selected as more relevant to a direction than more commonly used words such as `male' in these examples. This issue could be addressed in by determining a suitable prior for words but, as discussed above, we will leave that to future work.

\section{Effect of Computing Directions}
Our main paper describes PCA in CLIP-space as a way to generate candidate directions. However, the following alternative approaches were also considered. 

\paragraph{(i)~Random.} Random projections are a fast and popular feature transformation algorithm, especially for higher dimensional data. In Sec.~\ref{sec:ext_dir}, we further discuss some random directions extracted from the CLIP image space using the hybrid method.

\begin{figure}[h]
    \centering
    \includegraphics[width=\linewidth]{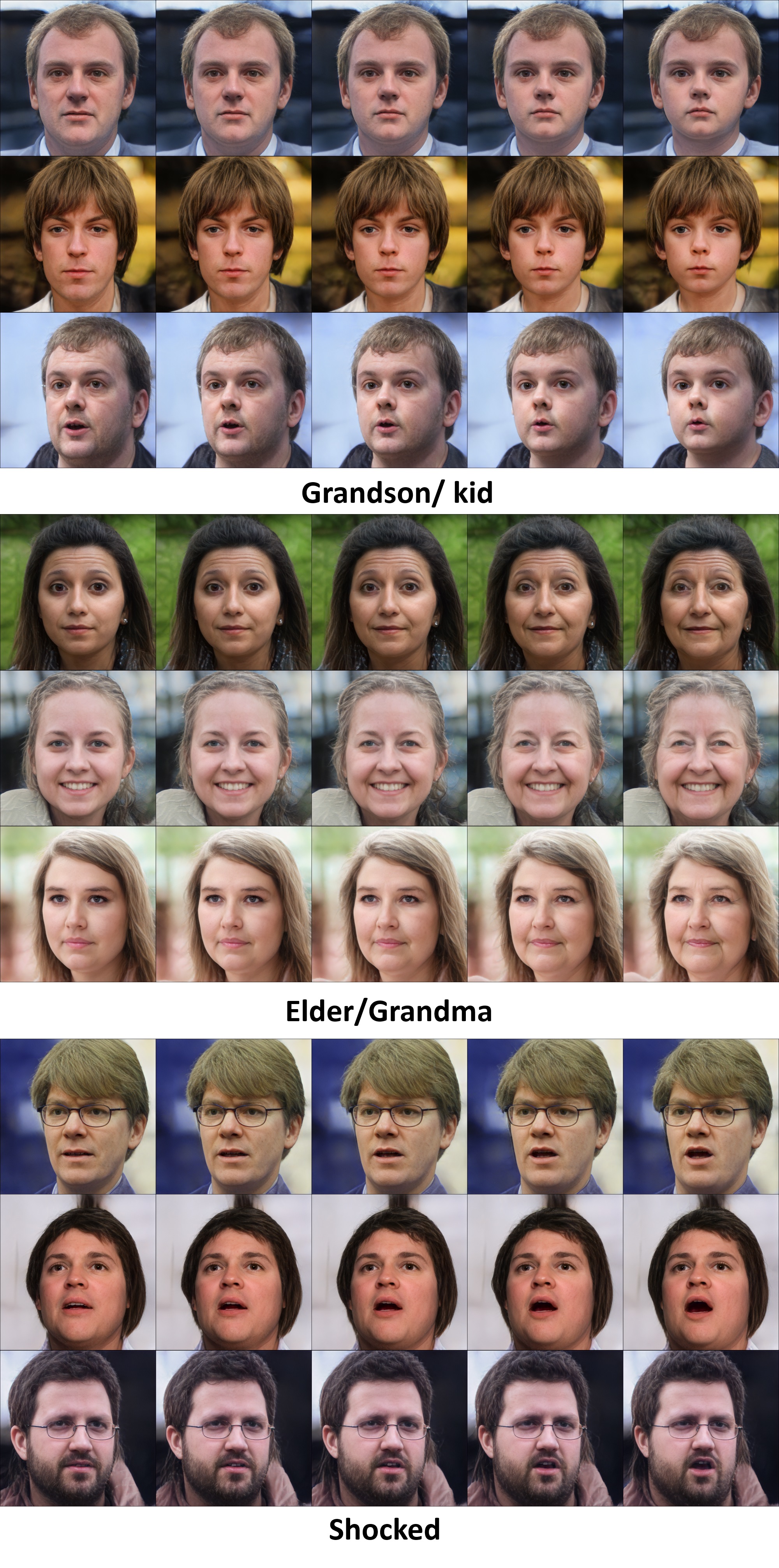}
    \caption{Some labeled edits extracted using the ICA analysis of the CLIP embeddings of FFHQ dataset and then projected to StyleGAN space. 
    }
    \label{fig:edits_ICA}
\end{figure}

\paragraph{(ii)~ICA.} Independent Component Analysis~(ICA) tries to extract independent directions that do not have mutual information. The output of ICA is also a set of $k$ direction vectors, each of length 512. An important difference between PCA and ICA is that PCA can compute all 512 directions even though the later directions, corresponding to smaller variance, may not be meaningful. In contrast, ICA only constructs $k \ll 512$ directions, e.g., 10 - 20 in most of our experiments. We extract multiple new directions from ICA analysis including ``Grandson/ kid", ``Elder/Grandma" and ``Surprised/shocked/excited" directions. Fig.~\ref{fig:edits_ICA} shows results of the CLIP2StyleGAN framework on the extracted directions using the ICA method.

\paragraph{Analysis.}
To analyze the consistency of PCA on the CLIP image embeddings, we use the direction extraction algorithm for a training dataset, e.g., FFHQ, and a StyleGAN generated dataset trained on the same training set.
The FFHQ dataset is expected to be more diverse than the corresponding StyleGAN generated image set.

PCA generally yields an ordered set of directions (discounting the effect of (near) duplicate duplicate eigenvalues). We can therefore compute the cosine similarity of the corresponding principal directions of the two datasets. We notice a decreasing trend in the similarity scores, \eg, the first principal direction of the StyleGAN distribution is aligned with the FFHQ distribution with a cosine similarity of $0.988$  and the last principal direction similarity reduces to $0.050$. This, empirically, indicates that StyleGAN is good at learning the important modes in the dataset but may not sufficiently capture the less dominant modes. This could also mean that the extraction of later PCA directions is noisy. In either case, the later principal directions of CLIP vectors in the FFHQ dataset \textit{cannot}  be used to do meaningful edits in the StyleGAN space, as they do not have a semantic meaning in the $W$ space of StyleGAN.

\begin{table}[h]
\caption{Labels of the entangled directions extracted and named, unsupervised, from the first 20 PCA directions using the \textbf{FFHQ} CLIP embeddings. Note that since we find some of these un-curated extracted directions to be offensive, we do not show the corresponding edit directions in the paper.  PC-$i$ denotes the $i$, most dominant, principal component. 
} 
\centering
\scriptsize
\begin{center}
    
\def\ua{\ensuremath{\uparrow}}
\def\da{\ensuremath{\downarrow}}

\begin{tabular}{rc} \toprule

 \multicolumn{1}{c}{PC}   & 
   \multicolumn{1}{c}{Labels}
 \\ \midrule
0  & Granddaughter, Kids, Kiddo, Children,Smile, Smiling \\
1  & Women, Female, Womeninstem, Womeninscience, Smile, Smiling\\
2  & Korean, Asian, Hye, Idol, Flawless, Wechat\\
3  & Asian, Chang\\
4  & Asian, Taiwanese, Malaysian, Developer, Filipino, Volunteer, Engineer\\
5  & Senior, Alum, Collegiate, Smiling, Mugshot \\
6  & Individual, Volunteer, Mexican, Lopez\\
7  & Arab, Latino, Israelis, Youth, Himalayan, Armenian\\
8  & Volunteer, Participant, Supporters, Missionary, Guides, Campaigning \\
9  & Youth, Young, Adolescent, Pupils, Turkish, Israeli, Bulgarian\\

10  & Journalist, Feminist, Traitor, Hypocrite, Women \\
11  & Medalist, Olympian, Young\\
12  & Jamaican, Blackgirl, Complexion, African, Michaela \\
13  & Individual, Member, Instructor\\
14  & Smile, Smiling, Pupils, Boy, Son, Smiley\\
15  &  Representative, Contestant, Professor, Scientist, Pharmacist \\
16  & Granddaughter, Pupils, Orphan, Youngsters, Youthful\\
17  & Trustee, Mentor, Advisor\\
18  &  Personnel, Cadets, Cap , Colonel, Officers \\
19  & Instructors, Coaches, Superintendent\\

\bottomrule
\end{tabular}
\end{center}
\label{tab: table_FFHQPCA20}
\end{table}

\begin{table}[h]
\caption{Labels of the entangled directions extracted and named, unsupervised, from the first 20 PCA directions using the \textbf{StyleGAN} CLIP embeddings. Note that since we find some of these un-curated extracted directions to be offensive, we do not show the corresponding edit directions in the paper.  PC-$i$ denotes the $i$, most dominant, principal component. 
} 
\centering
\scriptsize
\begin{center}
    
\def\ua{\ensuremath{\uparrow}}
\def\da{\ensuremath{\downarrow}}

\begin{tabular}{rc} \toprule

 \multicolumn{1}{c}{PC}   & 
   \multicolumn{1}{c}{Labels}
 \\ \midrule
0  & Niece, Child, Kiddo, Daughter, Smile \\
1  & Headshot, Womeninscience, Womenintech, Feminist\\
2  & Korean, Asian\\
3  & Taiwanese, Korean, Malaysian, Japanese, Programmer\\
4  & Malaysian, Filipino, Indonesian, Vietnamese, Nepali\\
5  &  Headshot, Granddaughter, Smiling, Student, Young\\
6  & Moroccan, Latino\\
7  & Teen, Youth, younger, Smiling, Mugshot\\
8  & Brokerage, Collaborator, Attorneys \\
9  & Volunteer, Student, Pupils, Youngster\\

10  &  Feminist, Journalist, Womeninscience\\
11  & Journalist, Writers, Individual\\
12  & Contestant, Commentator, Coordinator \\
13  & Volunteer, Builders\\
14  & Smile, Smiling, Youth\\
15  &  Comedians, Romanian, Norwegian, Swedish \\
16  & Smile, Contestant, Missionary, Presenter \\
17  & Chairman, Zucker, Manager, ***, Fraud, Man\\
18  &   Trustee, Volunteer, Psychologist, Councilor\\
19  & Scholar, Honorary, Instructor, Professor, Beard\\

\bottomrule
\end{tabular}
\end{center}
\label{tab: table_SGPCA20}
\end{table}
\begin{table}[t!]

\caption{ Disentangled directions and associated labels extracted from our PCA and hybrid analysis on faces. 
Here we list some of the disentangled directions corresponding to the ones shown in 
Table~\protect\ref{tab: table_FFHQPCA20} and 
\protect\ref{tab: table_SGPCA20}. 
Refer to the Sec.~\ref{sec: pretext} for the pretexts used. Note that a given direction can have multiple labels which we disentangle using our ``Refining Labels" section in the main paper. As notation, we use PC0a/PC0b to resepectively represent the disentangled directions for PC0, presented in the earlier tables.  
} 
\centering
\scriptsize
\begin{center}
    
\def\ua{\ensuremath{\uparrow}}
\def\da{\ensuremath{\downarrow}}

\begin{tabular}{rcc} \toprule

 \multicolumn{1}{c}{PC}   & 
  \multicolumn{1}{c}{Method} &   \multicolumn{1}{c}{Labels}
 \\ \midrule
0a & PCA (FFHQ/SG) & Granddaughter, Kids, Kiddo, Children \\
0b & PCA (FFHQ/SG) & Smile, Smiling\\
1a & PCA (FFHQ/SG) & Women, Female, Womeninstem, Womeninscience\\
1b & PCA (FFHQ/SG) & Smile, Smiling \\
19a & Hybrid (SG) & Beards, Beard, Grooming\\
19b & Hybrid (SG) & Glasses, Eyewear, Frames \\
35  & Hybrid (SG) & Cosplay, Cosmetics  \\
43 & Hybrid (SG) & ``Without" hair, ``No" hair , Bald \\

\bottomrule
\end{tabular}
\end{center}
\label{tab: table_dir_face}
\end{table}

\begin{table}[h]
\caption{ Some directions and associated labels extracted from our PCA and hybrid analysis on Cars. Refer to Sec.~\ref{sec: pretext} for the pretexts used. PC : Principal Component number. 
} 
\centering
\scriptsize
\begin{center}
    
\def\ua{\ensuremath{\uparrow}}
\def\da{\ensuremath{\downarrow}}

\begin{tabular}{rcc} \toprule

 \multicolumn{1}{c}{PC}   & 
  \multicolumn{1}{c}{Method} &   \multicolumn{1}{c}{Labels}
 \\ \midrule
0 & PCA & Burnout, Scrap, Charred, Rumble, Destroyed\\
1 & PCA & Capri, Camero, Classiccar\\
2 & PCA & Skoda, Escort\\
3 & PCA & Rocco, Subaru, Polo\\
12 & Hybrid & Blazer, SUV, Touring\\
44 & PCA & Red, Sedan, Mazda\\
79 & Hybrid & FD, Racecar, Sporty, Porsche\\
80 & Hybrid & Outlander, Volvo\\

\bottomrule
\end{tabular}
\end{center}
\label{tab: table_dir}
\end{table}

\section{Quality of the Extracted Directions}
\label{sec:ext_dir}
Figure 4 of the main paper shows directions extracted using the PCA and hybrid methods. Particularly, we extract ``Male/Female", ``Smile/No smile", ``Kids", \etc using PCA on the FFHQ dataset. Additionally, some other directions are extracted from the StyleGAN generated samples. Note that the directions extracted from these datasets using PCA can be similar, especially for the first few principal directions.Table~\ref{tab: table_FFHQPCA20} and Table~\ref{tab: table_SGPCA20} show top 20 directions extracted by the PCA method using the FFHQ and StyleGAN sampled CLIP embeddings. Note he we restrict more offensive labels but still show some predictions to assess the biases.

As discussed in Sec.~\ref{sec: impact}, the analysis also  extracts directions with a dominant ethnicity or profession label, we discard such directions due to ethical issues, and we therefore show an exclusive list of directions. The unsupervised extraction procedure in \name revealed some directions with inappropriate racial biases, and, given their offensive nature, we decided to \textit{not} show these directions. As a side note, we found \name to be quite effective at picking dominant directions present in the data, and thus reveal potential biases in the data that are otherwise hard to spot. This would an interesting future lead to collect (training) datasets that are less biased.

\begin{figure*}[t]
    \centering
    \includegraphics[width=\linewidth]{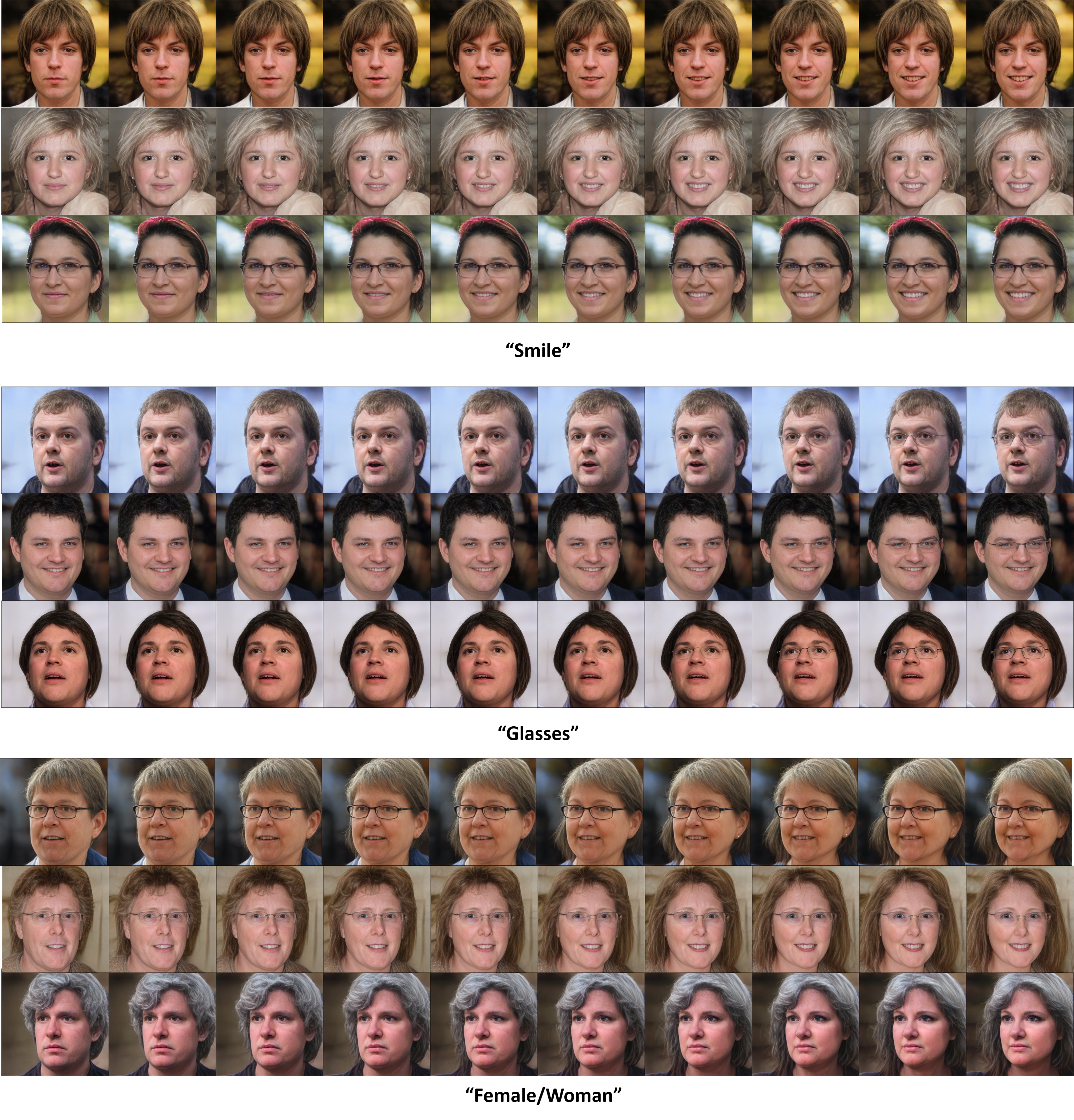}
    \caption{Some labeled edits extracted using the PCA/Hybrid analysis of the CLIP embeddings of faces and then projected to StyleGAN space. 
    }
    \label{fig:edits2}
\end{figure*}

\begin{figure*}[t]
    \centering
    \includegraphics[width=\linewidth]{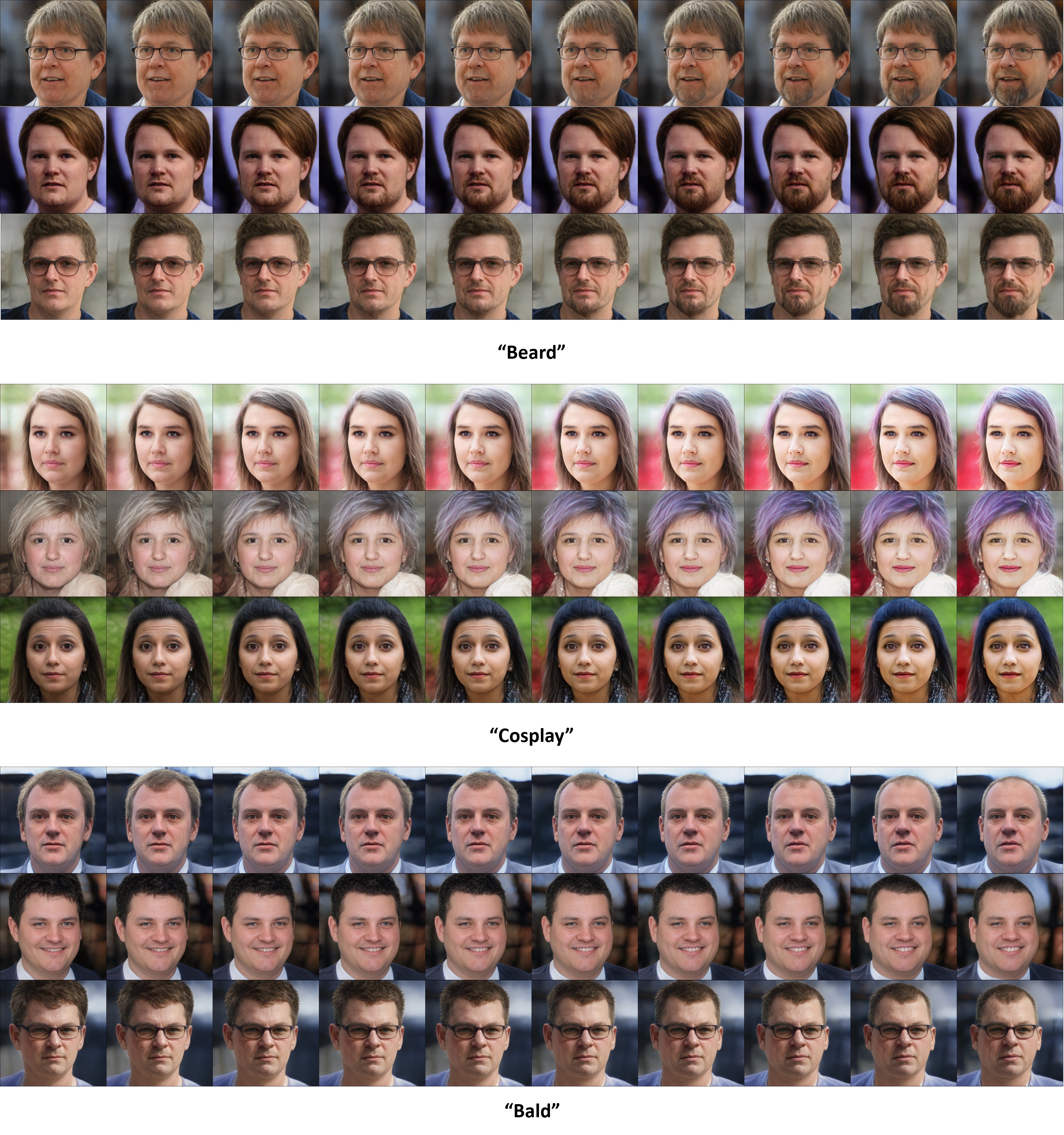}
    \caption{Some labeled edits extracted using the PCA/Hybrid analysis of the CLIP embeddings of faces and then projected to StyleGAN space. 
    }
    \label{fig:edits3}
\end{figure*}

\begin{figure*}[h]
    \centering
    \includegraphics[width=\linewidth]{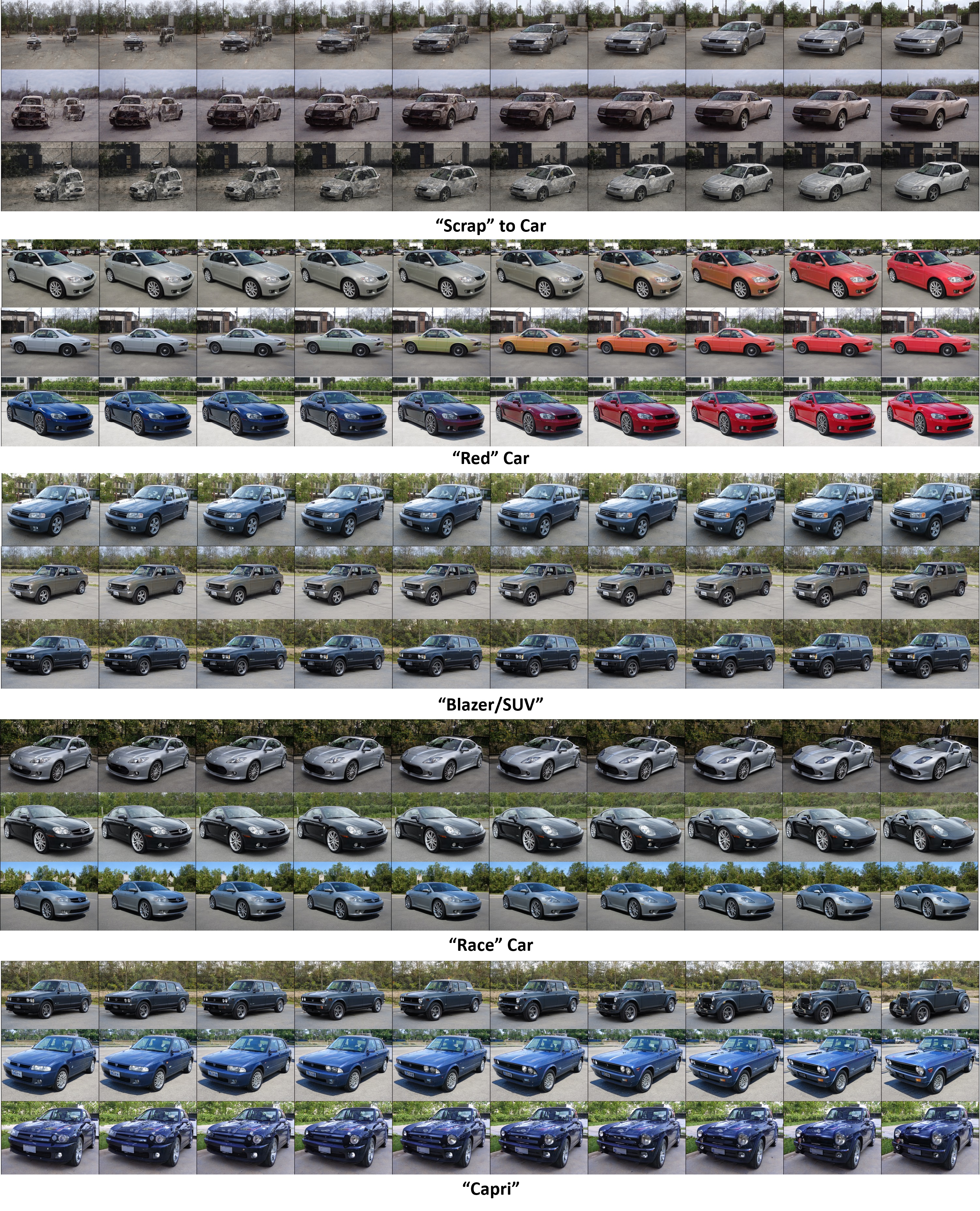}
    \caption{Some labeled edits extracted using the PCA/Hybrid analysis of the CLIP embeddings of cars and then projected to StyleGAN space. 
    }
    \label{fig:editscar}
\end{figure*}

For the LSUN-CAR dataset, we perform PCA on the StyleGAN generated samples to extract important directions. For example, the first principal direction using the PCA analysis on the CLIP image space is ``Scrap/Burnout/Charred".
This is a key example of the benefit of our approach, this is a meaningful and useful direction that, to the best of our knowledge, has not been used to edit car images in the past.
Table~\ref{tab: table_dir_face} and Table~\ref{tab: table_dir} show some directions and labels extracted using PCA and hybrid methods on the CLIP image embeddings which we visualize in this paper. Note that a given Principal component can be used to extract multiple directions using `` Refining Labels " section in the main paper. For instance in Table~\ref{tab: table_dir_face}, principal component number 0, 1, 19 have two directions associated with them.  Additionally, Fig.~\ref{fig:edits2}, Fig.~\ref{fig:edits3} and Fig.~\ref{fig:editscar} show some more results of edits on faces and cars.   

\section{List of Prompts}
\label{sec: pretext}
Here we describe the list of prefix text prompts used in Sec. 3 of the main paper. The prompts used are `A picture of a \{\}', `A picture of a \{\} person', `A picture of a \{\} car', `A picture of a person wearing \{\}', and  `A picture of a person with \{\}', `A picture of a person with \{\} hair', \etc.

\section{Identity Preservation}

\begin{table}[h]
\caption{ Identity preservation evaluation of various edits using state-of-the-art face recognition network~\cite{facerepo}. Note that the \textit{Glasses} and \textit{Cosplay} edits are not found by GANSpace.  Acc:Accuracy; CS:Cosine~Similarity.
} 
\centering
\scriptsize
\begin{tabular}{rrrrrrrr} \toprule

\multicolumn{1}{c}{Attribute}  & \multicolumn{1}{c}{Acc}  &\multicolumn{1}{c}{CS} &\multicolumn{1}{c}{Attribute}  & \multicolumn{1}{c}{Acc}  & \multicolumn{1}{c}{CS} \\ \midrule
Kids (Ours)  &0.90	&0.94 &  Male (Ours)  &0.80	&0.93 	 \\
Kids (GANSpace) &0.90 &0.94 & Male (GANSpace) &0.85 &0.93 \\
\midrule
Beard (Ours)  &1.00	&0.95 &  Smile (Ours)  &1.00	&0.97 	 \\

Beard (GANSpace)  &0.95 &0.94   & Smile (GANSpace)  &1.00	&0.98  \\
\midrule

Glasses (Ours)  &1.00	&0.98 &  Cosplay (Ours)  &1.00	&0.97	 \\

\bottomrule
\end{tabular}

\label{tab: table_identity}
\end{table}

\newcommand{\mcrot}[4]{\multicolumn{#1}{#2}{\rlap{\rotatebox{#3}{#4}~}}} 

\newcommand*{\twoelementtable}[3][l]%
{%
    \renewcommand{\arraystretch}{0.8}%
    \begin{tabular}[t]{@{}#1@{}}%
        #2\tabularnewline
        #3%
    \end{tabular}%
}

After we determine the accuracy of the edits applied to portrait images from FFQHQ in Sec. 4 of the main paper, in order to determine whether the edits performed in the $W^{+}$ space are identity preserving, we calculate the identity scores between the pairs of images along each direction.
We use the face recognition framework from dlib~\cite{facerepo}, which generates a unique descriptor for each face so that if the identity is the same in two images, the face-descriptors have small distance.   Table~\ref{tab: table_identity} shows the mean cosine similarity of the face embeddings between the original and the edited samples. We also report the average accuracy of the classifier for the pairs being the same person, using the distance tolerance value of $0.6$ to decide if the identity is preserved. Higher scores indicate that the edits are able to change the attributes without major changes in the identity of a person. Note that the ``Gender/Male" direction has a higher accuracy score for GANSpace but comparison with the Table 1 in the main paper indicates that the edit might retain some features or changes the attribute minimally which is not desirable.  Some edits including gender, if done successfully, should be expected to have an impact on identity.

\FloatBarrier
\clearpage
{\small
\bibliographystyle{ieee_fullname}
\bibliography{egbib}
}

\end{document}